\documentclass[10pt, twocolumn]{IEEEtran}

\usepackage{booktabs}    
\usepackage{graphicx}
\usepackage{adjustbox}
\usepackage{subfig}
\usepackage{adjustbox}
\usepackage{caption}  
\usepackage{multirow}    
\usepackage{tabularx}    
\usepackage{arydshln}    
\usepackage{graphicx}    
\usepackage{hyperref}    
\usepackage{lettrine}    
\usepackage{amsmath, amssymb}    
\usepackage{float}       
\usepackage{subcaption}  
\usepackage{textcomp}    
\usepackage{xcolor}      
\usepackage{cite}        
\usepackage{svg}         
\usepackage{siunitx}     

\title{EdgeAttNet: Towards Barb-Aware Filament Segmentation}

\author{
    Victor Solomon\textsuperscript{*}, 
    Piet Martens\textsuperscript{\dag}, 
    Jingyu Liu\textsuperscript{\dag}, 
    Rafal Angryk\textsuperscript{\ddag}\\
    
    Department of Computer Science, Georgia State University, Atlanta, GA, USA\\
    Email: vsolomon3@gsu.edu, rangryk@gsu.edu
}

\begin{document}

\maketitle
\begin{abstract}
Accurate segmentation of solar filaments in H-$\alpha$ observations is critical for determining filament chirality which is a key factor in determining the behavior of Coronal Mass Ejections (CMEs). However, existing methods often fail to capture fine scale filament structures, particularly barbs, due to a limited ability to capture long-range dependencies and spatial detail.

We propose EdgeAttNet, a segmentation architecture built on a U-Net backbone by introducing a novel, learnable edge map derived directly from the input image. This edge map is incorporated into the model by linearly transforming the attention Key and Query matrices with the edge information, thereby guiding the self-attention mechanism at the network’s bottleneck to more effectively capture filament boundaries and barbs. By explicitly integrating this structural prior into the attention computations, EdgeAttNet enhances spatial sensitivity and segmentation accuracy while reducing the number of trainable parameters.

Trained end-to-end, EdgeAttNet outperforms U-Net, and other U-Net-based transformer baselines on the MAGFILO dataset. It achieves higher segmentation accuracy and significantly better recognition of filament barbs, with faster inference performance which is suitable in practical segmentation model deployment. The full implementation and trained models are available at: \href{https://github.com/dasjar/EdgeAttNet}{\texttt{https://github.com/dasjar/EdgeAttNet}}.
\end{abstract}


\section{Introduction}

\begin{figure}[htbp]
    \centering
    \begin{minipage}[t]{0.48\columnwidth}
        \centering
        \includegraphics[width=\linewidth]{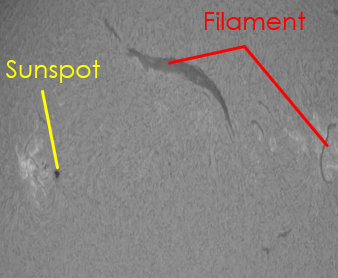}
        \\
        \textbf{(a)} Morphological differences between filaments and sunspots in an H-$\alpha$ observation.
        \label{fig:filamentvssunspot}
    \end{minipage}%
    \hfill
    \begin{minipage}[t]{0.48\columnwidth}
        \centering
        \includegraphics[width=\linewidth]{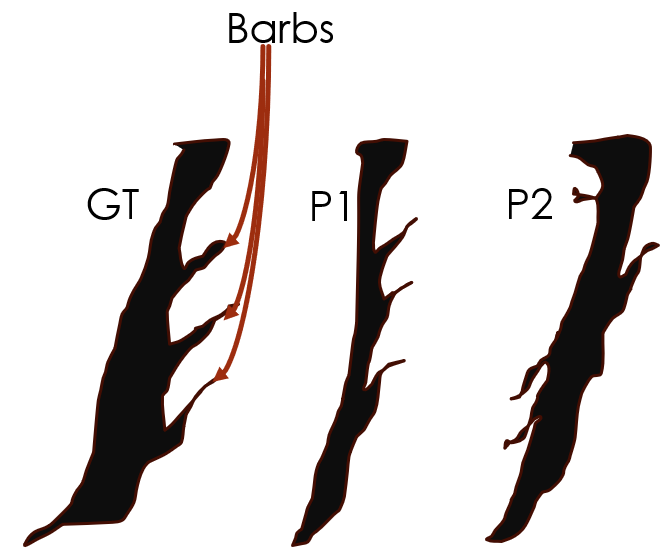}
        \\
        \textbf{(b)} Comparison of two predictions (P1 and P2) against the GT segmentation, highlighting barb orientation preservation.
        \label{fig:filament-bias}
    \end{minipage}
    \caption{Visual distinction between filaments and sunspots, and importance of capturing barbs during filament prediction in determining  chirality.}
    \label{fig:filament_and_prediction_comparison}
\end{figure}

\lettrine{F}{ilaments} are elongated, cool plasma structures suspended in the solar chromosphere. They can be visualized in H-$\alpha$ observations. They are mostly formed above polarity inversion lines in the corona\cite{gibson2018solar}. Filaments have similar intensity distribution with sunspots. Fig.~\ref{fig:filament_and_prediction_comparison}(a) illustrates the structural distinction between filaments and sunspots in a H-$\alpha$ observation.
Filaments are key precursors to solar eruptive events such as Coronal Mass Ejections (CMEs), which can severely disrupt satellite communications, navigation systems, and power grids \cite{eastwood2017scientific}. The destabilization and eruption of filaments is a major trigger for CMEs. Therefore, accurate filament segmentaion is relevant.

A crucial parameter in filament analysis is the filament's chirality which is a measure of the its magnetic field's orientaiton. The chirality influences the magnetic configuration and propagation behavior of associated CMEs\cite{martin1998conditions}. Determining filament chirality provides insight into the helicity and directionality of the erupting magnetic field, which are essential for modeling CME evolution and any potential eruption\cite{martin1998filament}.

An important visual cue for filament chirality identification is the orientation of its \textit{barbs}. Barbs are thin, lateral protrusions extending from the filament's spine. The angle and direction of a filament's barbs relative to its spine indicates the filament's chirality \cite{hao2016can}. Accurate capturing of barbs during filament segmentation is therefore considered relevant for reliable chirality identification. Fig.~\ref{fig:filament_and_prediction_comparison}(b) highlights the importance of accurately capturing the barbs of a filament. Two predicted masks (P1 and P2) with the Ground Truth (GT) segmentation are shown. While P2 demonstrates a higher overall overlap with the GT, it fails to capture essential barb features. In contrast, P1, despite exhibiting a lower overall overlap more accurately preserves the barb orientation, thereby enabling more precise and reliable chirality determination.

\begin{figure}
    \centering
    \includegraphics[width=1.0\linewidth]{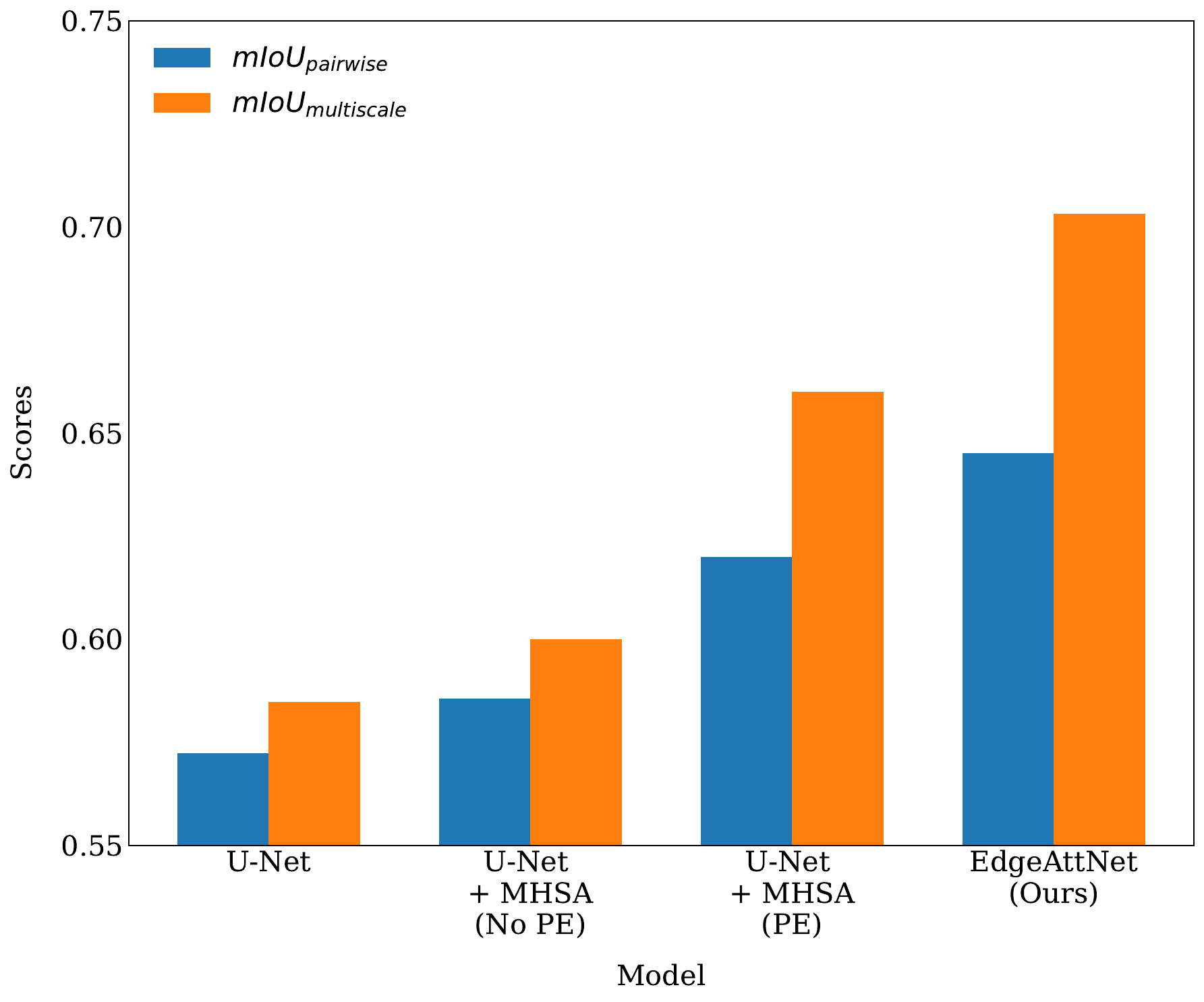}
    \caption{Performance of EdgeAttNet vs. baselines. Evaluation is conducted on the MAGFILO test split using the $\mathit{mIoU}_{\mathit{pairwise}}$ and $\mathit{mIoU}_{\mathit{multiscale}}$ metrics, which are discussed in Section III of this study. EdgeAttNet achieves the best performance across all metrics.}

    \label{radarplot}
\end{figure}

Studies such as \cite{ahmadzadeh2019toward, zhu2025flat} employ convolutional layers and channel attention mechanisms respectively for filament segmentation. However, convolutional networks have a limited receptive field and therefore struggle to capture global context. Additionally, the attention mechanisms used in these works often require a large number of trainable parameters due to the inclusion of positional encodings and lack proper attention guidance that shifts the attention to better capture filament boundaries and barbs. In contrast, our method eliminates the need for positional encodings (PEs) by introducing a learned edge map derived directly from the input image. This edge map guides the attention mechanism while sufficing for the absence of PEs thereby improving segmentation and improved recognition of barbs whilst significantly reducing the number of trainable parameters in the model as compared to SOTA and baseline models. Fig.~\ref{radarplot} shows that EdgeAttNet outperforms SOTA benchmarks and baseline models across all represented evaluation metrics.

The remainder of this paper is organized as follows: Section II discusses existing works that are related to our study and their limitations. Section III describes the acquisition and preprocessing of the data used in our work. Section IV discusses the evaluation metrics used in evaluating performance of models. Section V introduces our proposed method. Section VI outlines the experimental setup and the results of our work. Section VII concludes with key findings and directions for future work.

\section{Related Work}
In this section, we review existing segmentation models, highlight their limitations, and explain how our proposed model addresses these challenges.

Architectures such as U-Net~\cite{ronneberger2015u} and U-Net Transformer~\cite{petit2021u} have demonstrated strong performance in the segmentation domain but exhibit several limitations. U-Net~\cite{ronneberger2015u} struggles to preserve long-range context due to its limited receptive field. In contrast, U-Net Transformer~\cite{petit2021u} incorporates multiple Multi-Head Self-Attention (MHSA) modules across several layers, each relying on positional encodings (PEs), which significantly increases the number of trainable parameters. Moreover, these MHSA modules are not explicitly guided to attend to edge structures in the image, which may limit their ability to accurately capture fine-grained features such as boundaries and barbs.

\subsection{U-Net and Variants}
U-Net~\cite{ronneberger2015u} uses an encoder–decoder structure with skip connections from corresponding encoder layers to decoder layers, which help preserve spatial resolution. For this reason, we use it as a baseline in our work. However, its reliance on fixed receptive fields limits its ability to capture long-range contextual dependencies~\cite{qin2020match}.

Attention U-Net~\cite{oktay2018attention} extends the original U-Net by introducing attention gates that enhance salient features and suppress irrelevant background responses, particularly within skip connections. While effective in biomedical segmentation, these gates rely on gating signals from coarser scales and do not explicitly model long-range dependencies or global context. This can limit their effectiveness in particularly in filament segmentation, where capturing the barbs is important.

\subsection{Vision Transformers in Segmentation}
Transformer-based models, such as SegFormer \cite{xie2021segformer}, have demonstrated impressive performance in semantic segmentation by leveraging hierarchical encoders, patch embeddings, and efficient attention mechanisms. SegFormer employs a lightweight Multi-Layer Perceptron (MLP) decoder and omits traditional positional encodings, achieving strong generalization across standard benchmarks. However, the absence of explicit edge guidance may limit the model's ability to capture fine edge details.

\begin{figure*}[htbp]
    \centering
    \includegraphics[width=\textwidth]{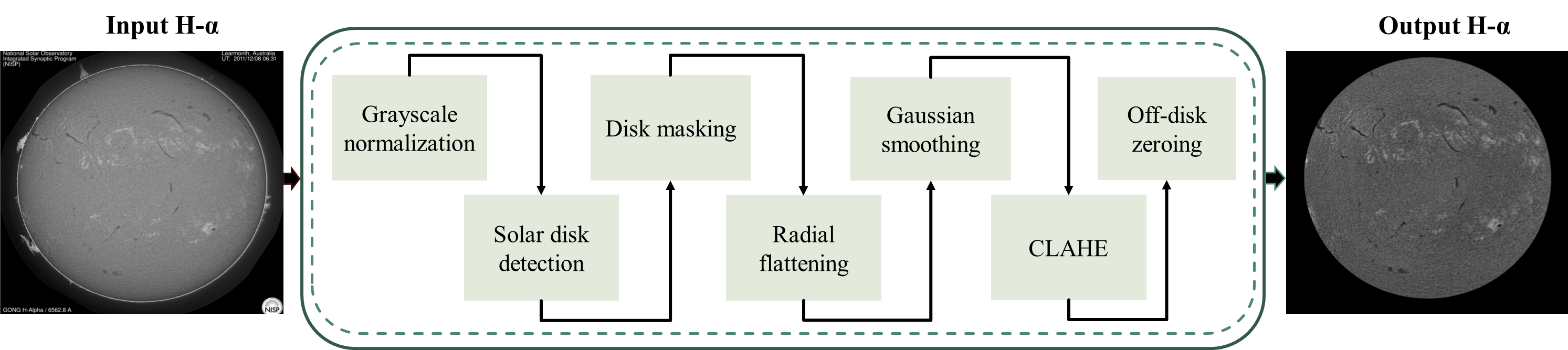}
    \caption{Preprocessing pipeline for the 1,593 H-$\alpha$ observations obtained from GONG. Each image is temporally aligned with its corresponding annotation in the MAGFILO dataset. The preprocessed data is available for use at: \href{https://github.com/dasjar/EdgeAttNet}{\texttt{https://github.com/dasjar/EdgeAttNet}} }
    \label{processing}
\end{figure*}

Attention mechanisms such as CBAM~\cite{woo2018cbam}, Squeeze-and-Excitation (SE)~\cite{hu2018squeeze}, and Spatial and Channel Squeeze-and-Excitation (SCSE)~\cite{roy2018concurrent} enhance feature learning by emphasizing spatially or channel-wise informative regions. However, these methods do not incorporate explicit attention guidance, which may limit their capacity to focus on fine structural details such as edges in certain segmentation tasks.

More recently, models like EdgeFormer \cite{ge2022edgeformer} have embedded edge cues into the attention mechanism to enhance boundary localization. While effective for natural image tasks, such approaches may face challenges in generalizing to scientific imaging domains such as filament segmentation where preserving thin, low-contrast structures is critical.

\subsection{Addressing Existing Limitations}

To address these challenges, our proposed method incorporates edge maps learned directly from the input image and integrates them with the attention mechanism solely at the bottleneck of the U-Net. This guides the attention toward capturing edges and barbs in the image whilst reducing model complexity. The combination of the learned edge map with Multi-Head Self-Attention (MHSA) forms the core of our Edge-Guided Multi-Head Self-Attention (EG-MHSA) module, effectively compensating for the absence of PEs while directing attention toward structurally relevant features.

\subsection{Contributions}
The main contributions of our work are summarized as follows:

\begin{itemize}
    \item A robust data processing pipeline for H-$\alpha$ observations that can be adapted to any H-$\alpha$ dataset to improve model training performance.
    \item We propose our novel segmenation method that uses learned edge maps from the input image and use them to make a linear transformation of the attention mechanism to guide attention in a novel manner thereby shifting the attention of the  model to recognize boundaries and barbs of the input object.
    \item Evaluation of our proposed method and other SOTA and baseline models on the MAGFILO\cite{ahmadzadeh2024dataset} dataset.
\end{itemize}

\section{Data Acquisition and Preprocessing}

In this section, we discuss the data used for this study and the preprocessing we have applied to prepare it for model training.

H$\alpha$ observations, observed at a wavelength of \SI{6562.8}{\angstrom}, were obtained from the Global Oscillation Network Group (GONG)~\cite{harvey1996global}, which operates six ground-based observatories providing continuous 24-hour solar monitoring. Each station captures observations at six-minute intervals, yielding a combined global cadence of approximately one image per minute.

\begin{figure*}[htbp]
    \centering
    \includegraphics[width=\textwidth]{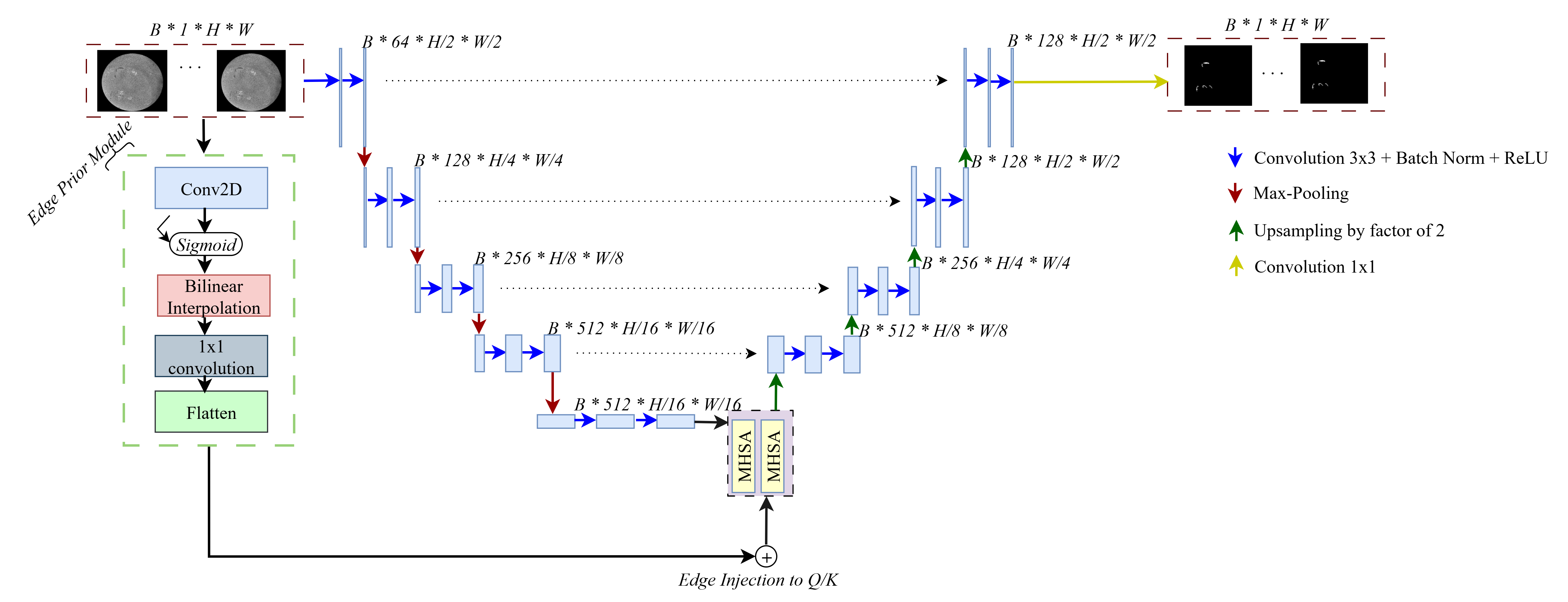}
    \caption{Architecture of our proposed EdgeAttNet model. It combines the edge prior with two EG-MHSA at the bottleneck of the U-Net. The model takes a grayscale H-$\alpha$ image as input and outputs a binary segmentation mask.}
    \label{architecture}
\end{figure*}

This high temporal resolution yields a large volume of data, making it well-suited for data-driven methods such as deep learning. However, deep learning models require well-annotated data for effective supervised training. In solar physics, annotated data remains scarce, presenting a significant challenge~\cite{wang2014new}.

To address this limitation, we utilize the Manually Annotated GONG Filaments in H$\alpha$ Observations (MAGFILO) data~\cite{ahmadzadeh2024dataset}, which provides manually labeled masks for filaments. MAGFILO follows the widely adopted COCO data style annotation format, which includes bounding boxes, polygons, and filament spine annotations for 1,593 H$\alpha$ observations collected between 2011 and 2022. Of these, 958 observations are unique~\cite{ahmadzadeh2024dataset}. Notably, all annotations were curated manually, and in some cases, multiple annotators may have annotated the same observation. While this ensures a high level of detail, it is important to acknowledge that perfectly annotating complex solar phenomena, such as filaments, is nearly impossible. Consequently, the MAGFILO dataset may still exhibit certain limitations.

\subsection{Data Preprocessing}

While H$\alpha$ observations are well-suited for visualizing filaments, they often contain inherent artifacts that can introduce noise in deep learning applications. Common issues include limb darkening and interference from unrelated chromospheric fibrils.

To mitigate these issues, we curated a set of 1,593 GONG H$\alpha$ observations, each precisely time-aligned with its corresponding annotation from the MAGFILO dataset. The associated FITS (Flexible Image Transport System) files were retrieved using the SunPy library and converted to JPEG format for standardized preprocessing.

Each image passes through our preprocessing pipeline as shown in Fig.~\ref{processing}, which is designed to enhance filament visibility and suppress noise. The pipeline includes normalization, disk masking, radial flattening~\cite{men2017background}, Gaussian smoothing~\cite{wink2004denoising}, and contrast enhancement~\cite{reza2004realization}.

The pipeline begins by normalizing pixel intensities to the $[0, 1]$ range, ensuring numerical stability in subsequent steps. The solar disk is then localized using the Hough Circle Transform\cite{duda1972use}, producing accurate estimates of its center and radius. These parameters are used to create a binary mask that excludes off-disk regions, ensuring that all further processing is confined to the solar disk.

To correct for radial brightness gradients caused by limb darkening, we implement a radial flattening technique~\cite{men2017background}. This involves calculating the normalized radial distance of each pixel from the disk center and aggregating pixel intensities into concentric annular bins. A smoothed 1-D profile is then interpolated into a 2D background estimate used to normalize the image. This operation equalizes luminance across the disk while preserving filament structures.

The background-corrected image is then smoothed using a Gaussian blur~\cite{wink2004denoising} with a $3 \times 3$ kernel and a standard deviation of $\sigma = 0.7$, which suppresses high-frequency noise while retaining edge details. Finally, Contrast Limited Adaptive Histogram Equalization (CLAHE)~\cite{reza2004realization} is applied within the disk mask. CLAHE enhances local contrast and amplifies faint filamentary signals, especially in low-intensity regions, while preventing over-enhancement in brighter areas. All off-disk pixels are explicitly set to zero to ensure spatial consistency.

The final output is a set of normalized, contrast-enhanced, and spatially masked H$\alpha$ observations with significantly improved signal-to-noise ratios and clearer filament structures, as illustrated in Fig.~\ref{processing}. These processed observations constitute a high-quality, model-ready dataset suitable for downstream tasks such as filament segmentation using deep neural networks.

\section{Evaluation Metrics}

Previous studies, such as~\cite{guo2022solar} and~\cite{diercke2024universal}, have evaluated filament segmentation models using standard region-based accuracy metrics. However, in our work, we place emphasis on accurately capturing the edges and barbs of filaments, as these features are essential for chirality identification~\cite{Ji2023}, as mentioned in Section~I of this study. This motivated us to use a broader set of evaluation measures.

Traditional metrics such as average precision (AP) and average recall (AR), while widely used in object detection benchmarks like the Common Objects in Context (COCO) dataset~\cite{lin2014microsoft}, are not specifically designed to evaluate how well filament barbs are captured, as pointed out in Fig.~\ref{fig:filament_and_prediction_comparison}(b). To address this limitation, we employ evaluation metrics that more accurately assess segmentation quality based on both region overlapping and edge recognition.




\subsection{Pairwise IoU ($\text{IoU}_{\text{pairwise}}$)}

To assess segmentation accuracy at the level of individual filament instances, we perform a pairwise IoU\cite{rezatofighi2019generalized} evaluation for each image. For a given observation, all pairs $(gt_i, pt_j)$ with non-zero spatial intersection are evaluated:
\begin{equation}
\mathit{IoU}_{\mathit{pairwise}}(gt_i, pt_j) = 
\begin{cases}
\frac{|gt_i \cap pt_j|}{|gt_i \cup pt_j|}, & \text{if } |gt_i \cap pt_j| > 0 \\
0, & \text{otherwise}
\end{cases}
\end{equation}

This metric isolates spatially relevant matches, removing the influence of missed or spurious detections. It is especially informative when segmentations involve object splits or merges. However, it may penalize correct detections that differ only in granularity, such as a single ground-truth filament segmented into multiple detected parts.



\subsection{Multiscale IoU (\textit{IoU}\textsubscript{multiscale})}
To assess the structural accuracy of filament segmentation, particularly in correctly capturing barbs, we adopt the multiscale Intersection over Union ($\mathit{IoU}_{\mathit{multiscale}}$) metric~\cite{ahmadzadeh2021multiscale}. Unlike standard $\mathit{IoU}$, which operates at a single resolution, $\mathit{IoU}_{\mathit{multiscale}}$ evaluates overlap across multiple spatial scales, making it sensitive to missing the edges of objects.

Given a ground-truth object $o$ and a predicted region $\tilde{o}$, masks are downsampled at multiple resolutions $\delta_i \in \Delta$ using a function $s(o, \delta_i)$. The intersection ratio at each scale is:
\begin{equation}
r(o, \tilde{o}, \delta_i) = \frac{n\left( s(o, \delta_i) \cap s(\tilde{o}, \delta_i) \right)}{n\left( s(o, \delta_i) \right)}
\end{equation}
where $n(\cdot)$ counts non-zero grid cells.

The final mIoU is the average over scales:
\begin{equation}
\mathit{IoU}_{\mathit{multiscale}}(o, \tilde{o}) = \int_{0}^{1} r(o, \tilde{o}, \delta) \, d\delta
\end{equation}
This is approximated as a discrete sum over a set of scales $\Delta$, capturing both fine and coarse structures.

$\mathit{IoU}_{\mathit{multiscale}}$ emphasizes boundary alignment and topological fidelity, making it particularly effective for our segmentation task. Combined with \textit{IoU}\textsubscript{pairwise}, We aim to attain a robust evaluation.

\begin{table*}[t]
\centering
\caption{Architectural comparison between U-Net variants and our proposed EdgeAttNet. "PE" refers to positional encoding. EdgeAttNet replaces positional embeddings with edge-guided attention for improved structural bias and parameter efficiency.}
\renewcommand{\arraystretch}{1.3}
\setlength{\tabcolsep}{6pt}
{\fontsize{9}{11}\selectfont
\begin{tabularx}{\textwidth}{l|X|X|X|X}
\hline
\multirow{2}{*}{\textbf{Characteristic}} & \multicolumn{3}{c|}{\textbf{U-Net Variants}} & \textbf{EdgeAttNet (Ours)} \\
\cdashline{2-4}
& U-Net & U-Net + MHSA (No PE) & U-Net + MHSA (PE) & \\
\hline
Architecture Type & CNN Encoder–Decoder & + MHSA & + MHSA (PE) & Hybrid CNN–Transformer with EG-MHSA \\
\hline  

Backbone Depth & 4-stage Conv & 4-stage Conv & 4-stage Conv & 4-stage Conv + 2 EG-MHSA Blocks \\

\hline  
Decoder Type & Symmetric Upsampling & Symmetric Upsampling & Symmetric Upsampling & Upsampling + Edge-Aware Decoder \\
\hline  
Pretrained Weights & None & None & None & None \\
\hline  
Attention Mechanism & None & MHSA (No PE) & MHSA (PE) & EG-MHSA \\
\hline  
Number of Parameters & 31,030,593 & 35,231,041 & 35,362,113 & 22,658,891 \\
\hline
\end{tabularx}
}
\label{architecture-comparison}
\end{table*}

\section{Methodology}

In this section, we describe the architecture of our proposed model, EdgeAttNet. We outline its overall structure and elaborate on key components, including the edge map extraction, the edge-guided multi-head self-attention (EG-MHSA) mechanism, and our strategy for integrating the edge map at the bottleneck of the U-Net. We also discuss the loss functions used in training the model.

\subsection{Overall Architecture}
As shown in Fig.~\ref{architecture}, EdgeAttNet follows a U-Net-style encoder-decoder architecture enhanced with edge-aware attention. Let the input grayscale image be denoted by $\chi \in \mathbb{R}^{B \times 1 \times H \times W}$, where $B$ is the batch size and $H \times W$ are the height and width of the image respectively. The corresponding ground truth segmentation mask is denoted as $y \in \{0,1\}^{B \times 1 \times H \times W}$.

The encoder consists of four hierarchical stages, each comprising a double convolution block followed by a $2\times2$ max-pooling layer. Each double convolution includes two sequential $3\times3$ convolutions with batch normalization and ReLU activation. In our work, the number of feature channels increases as $[64, 128, 256, 512]$, thereby reducing spatial resolution by a factor of 16 at the bottleneck.

After encoding, the feature map passes through a bottleneck composed of a double convolution block followed by two stacked EG-MHSA modules. These modules inject edge-guided attention into the network’s deepest representation, promoting a boundary-aware global context modeling. The decoder path mirrors the encoder with transposed convolutions for upsampling, skip connections, and double convolution blocks for feature fusion. A final $1 \times 1$ convolution outputs the segmentation logits.

\subsection{Encoder and Decoder Path}
The encoder gradually contracts the spatial resolution while increasing feature richness. At each stage, we preserve intermediate feature maps for skip connections. The decoder path progressively restores spatial resolution using bilinear upsampling via transposed convolutions, concatenates the corresponding encoder features, and refines them using double convolution blocks. This symmetric design facilitates detailed reconstruction of spatial structures from coarse representations. 

Formally, if $x_i$ denotes the output of the $i^{\text{th}}$ encoder stage, then the decoder reconstructs from the bottleneck output $z$ as $\hat{x}_i = \text{DoubleConv}(\text{Concat}(x_i, \text{UpConv}(z)))$, where $\text{UpConv}$ denotes a $2 \times 2$
transposed convolution.

\subsection{Edge Prior Extraction}
To enhance spatial sensitivity, we introduce a lightweight edge prior branch that predicts edge-like features directly from the input image. This branch comprises a single $3 \times 3$ convolution followed by a sigmoid activation, resulting in an edge map $E = \sigma(\text{Conv}_{3 \times 3}(\chi)) \in \mathbb{R}^{B \times 1 \times H \times W}$.

The edge map $E$ is then bilinearly interpolated to match the spatial resolution of the bottleneck feature map of $H/16 \times W/16$. It is further projected to the corresponding channel dimension via a $1 \times 1$ convolution, yielding:\[
E' = \text{Conv}_{1 \times 1}(\text{Interpolate}(E)) \in \mathbb{R}^{B \times C \times H' \times W'}
\]
This transformed edge bias $E'$ is added to the query and key embeddings within the attention mechanism. It serves as a structural guide, emphasizing boundaries and suppressing irrelevant spatial regions.

\begin{table}[ht]
\centering
\caption{Comparison of our proposed EdgeAttNet model and baseline models tested on the MAGFILO test split. EdgeAttNet shows superior performance.}
\renewcommand{\arraystretch}{1.3}
\setlength{\tabcolsep}{2.5pt}
{\footnotesize
\begin{tabular}{l|p{1.15cm}|p{1.25cm}|p{1.25cm}|p{1.25cm}}
\hline
\textbf{Metric} & \multicolumn{3}{c|}{\textbf{U-Net Variants}} & \textbf{EdgeAttNet (Ours)} \\
\cdashline{2-4}
& \centering U-Net & \centering U-Net \\ + MHSA \\ (No PE) & \centering U-Net \\ + MHSA \\ (PE) & \\
\hline
$\mathit{mIoU}_\mathit{pairwise}$   & 0.5724 & 0.5856 & 0.6200 & \textbf{0.6451} \\
\hline
$\mathit{mIoU}_\mathit{multiscale}$ & 0.5848 & 0.6000 & 0.6601 & \textbf{0.7032} \\
\hline
\end{tabular}
}
\label{scoretable}
\end{table}

\subsection{Edge-Guided Multi-Head Self-Attention (EG-MHSA)}
We augment the standard multi-head self-attention mechanism with a learned edge prior to enhance boundary sensitivity. This attention module operates on feature maps that are flattened along the spatial dimensions. Let $x_i$ denote the output feature map from the $i^{\text{th}}$ encoder stage. Specifically, we use $x_n$ to represent the bottleneck feature map, which serves as the input to the EG-MHSA module:
\[
x_n \in \mathbb{R}^{B \times C \times H' \times W'} \rightarrow x_{n,\text{flat}} \in \mathbb{R}^{B \times (H'W') \times C}.
\]

\begin{figure*}[t!]

\centering
\hspace*{-5.0em}  
\includegraphics[width=0.7\textwidth]{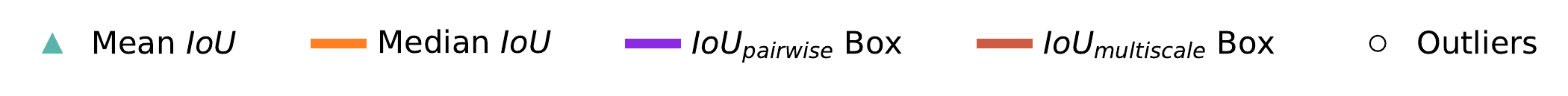}

    \vspace{-0.2em}

    \centering

    \begin{minipage}[c]{0.06\textwidth}
        \includegraphics[width=\linewidth]{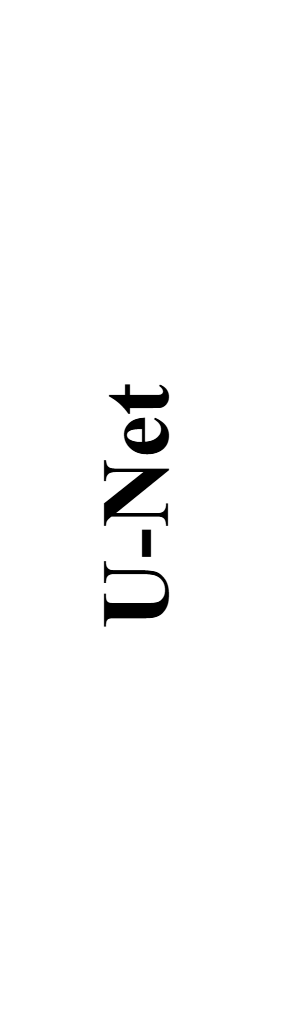}
    \end{minipage}%
    \hspace{-1.5em}  
    \vspace{1.8em}
    \begin{minipage}[c]{0.9\textwidth}
        \includegraphics[width=\linewidth]{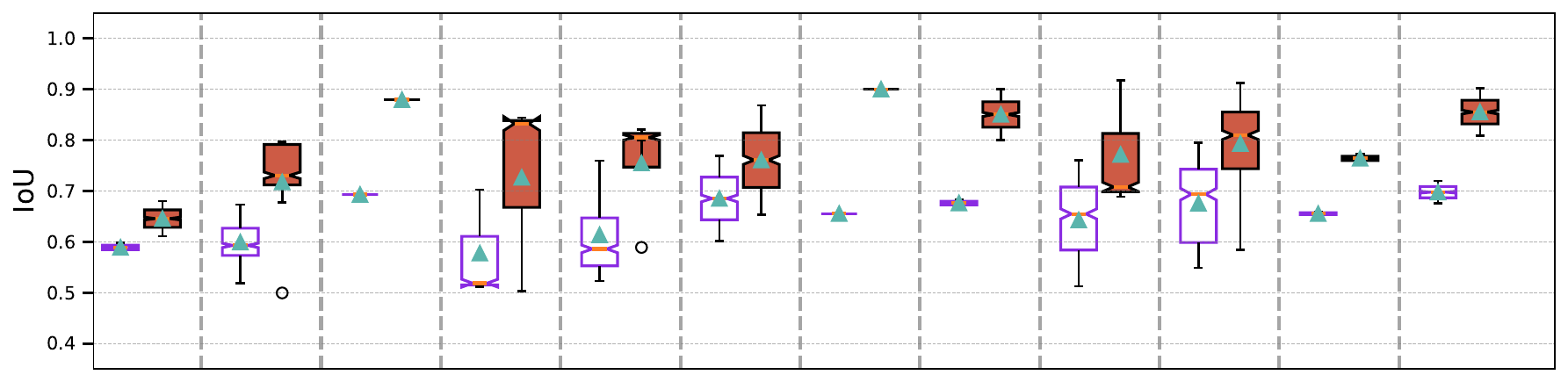}
    \end{minipage}

        \vspace{-1.5em}

    \begin{minipage}[c]{0.063\textwidth}
        \includegraphics[width=\linewidth]{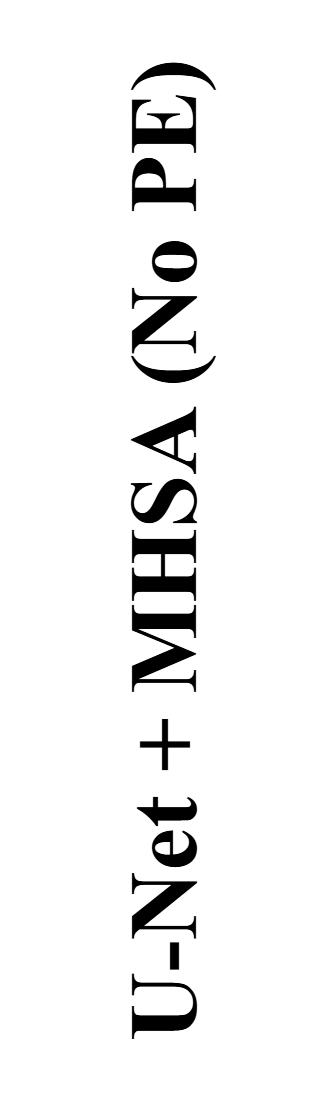}
    \end{minipage}%
    \hspace{-1.5em}
    \vspace{1.8em}
    \begin{minipage}[c]{0.9\textwidth}
        \includegraphics[width=\linewidth]{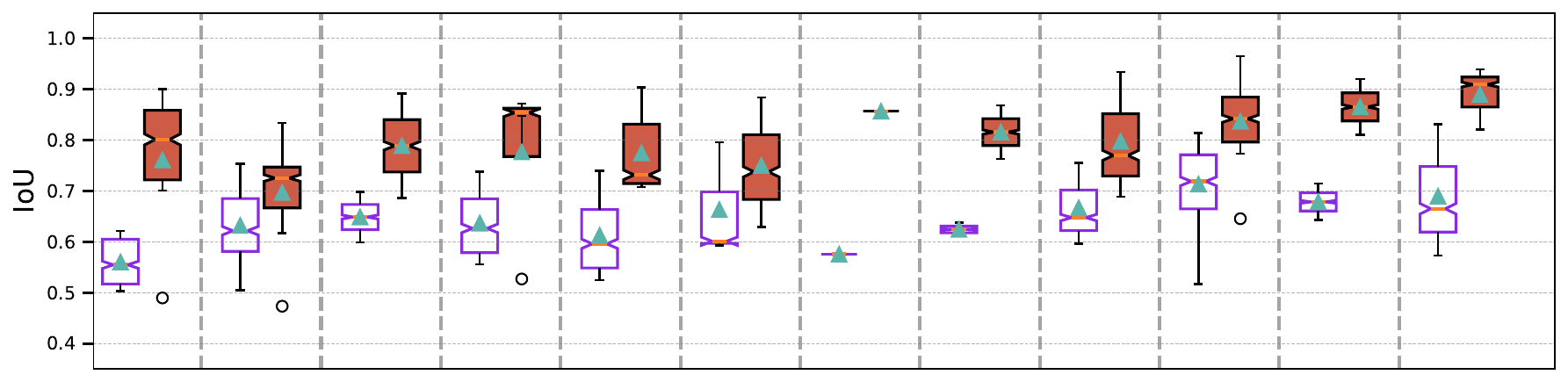}
    \end{minipage}

    \vspace{-1.5em}

    \begin{minipage}[c]{0.06\textwidth}
        \includegraphics[width=\linewidth]{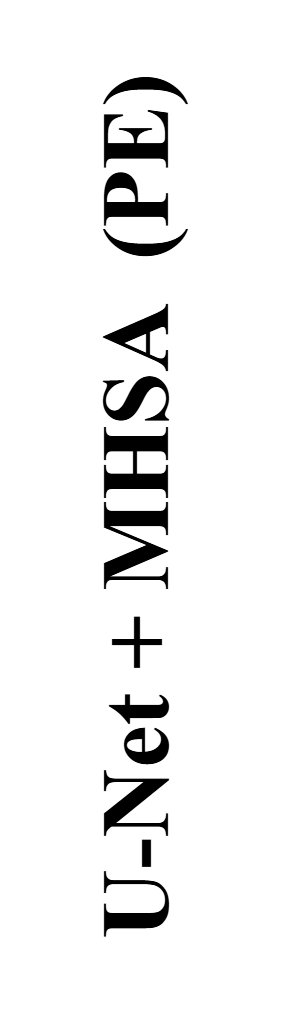}
    \end{minipage}%
    \hspace{-1.5em}
    \vspace{1.8em}
    \begin{minipage}[c]{0.9\textwidth}
        \includegraphics[width=\linewidth]{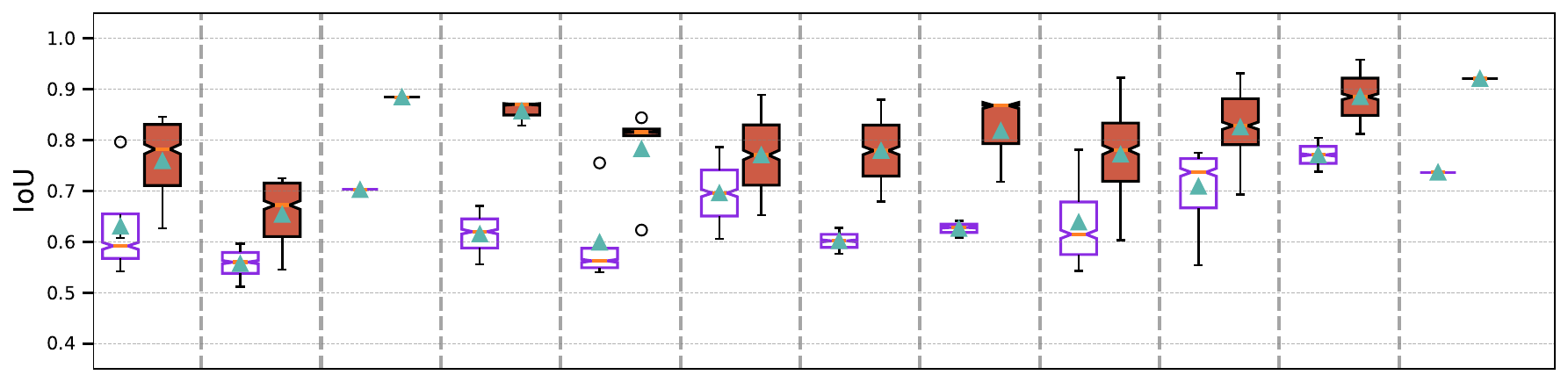}
    \end{minipage}

     \vspace{-1.5em}

\raisebox{4.0em}{ 
  \begin{minipage}[c]{0.07\textwidth}
    \hspace*{-1.1em}
    \includegraphics[width=\linewidth]{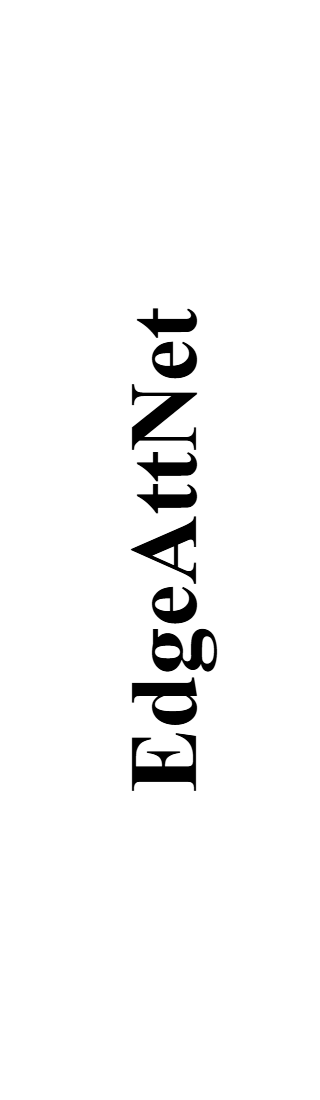}
  \end{minipage}%
}%
\hspace{-2.2em}
\begin{minipage}[c]{0.9\textwidth}
  \includegraphics[width=\linewidth]{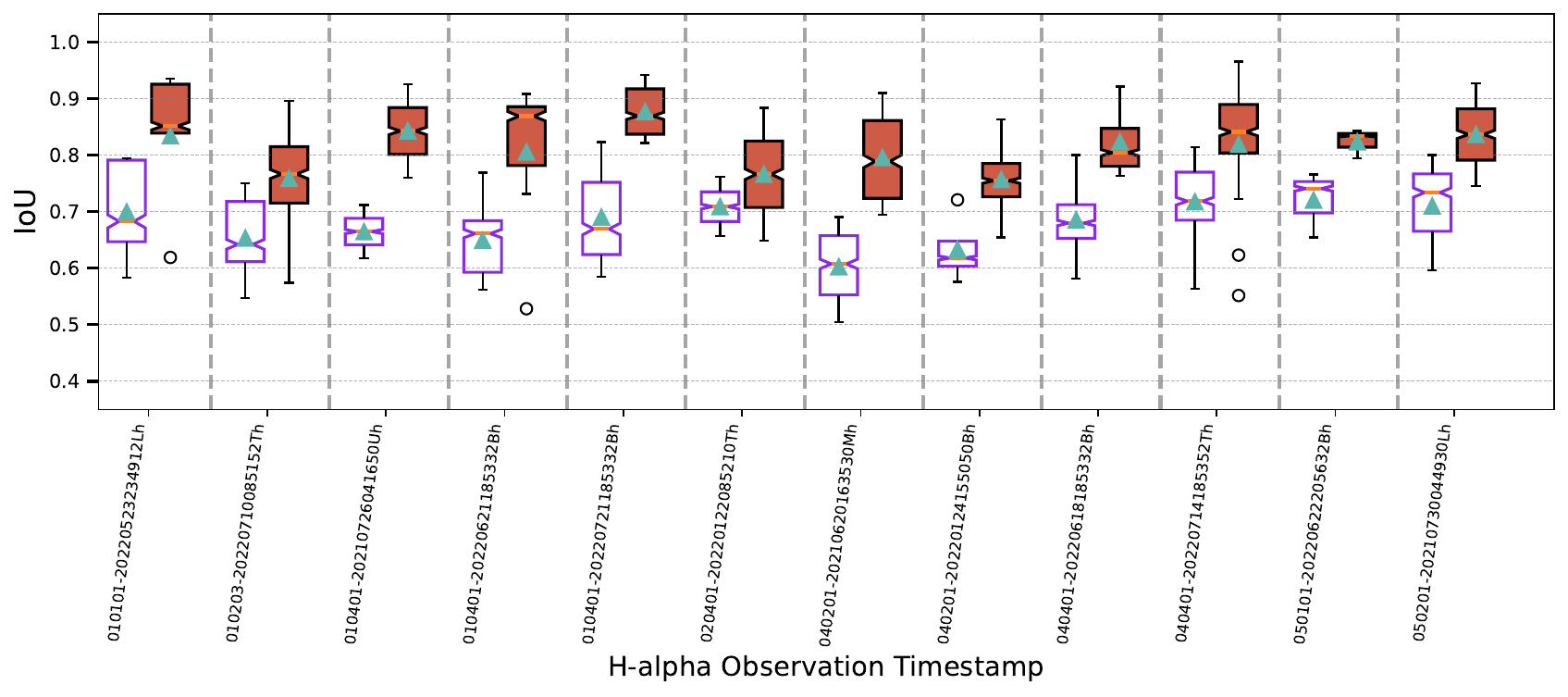}
\end{minipage}

    \caption{Comparison of segmentation performance across 12 randomly selected H$\alpha$ from our MAGFILO test split. Each row shows the boxplots for a specific model. Each boxplot represents the distribution of $\mathit{IoU}_{\mathit{pairwise}}$ scores for that sample.}
    \label{boxplots}

\end{figure*}

The edge prior, denoted as $E' \in \mathbb{R}^{B \times (H'W') \times C}$, is added to the input features to guide the computation of attention. The query ($Q$), key ($K$), and value ($V$) matrices are constructed as:
\[
Q = x_{n,\text{flat}} + E', \quad K = x_{n,\text{flat}} + E', \quad V = x_{n,\text{flat}}.
\]

Self-attention is then computed as:
\[
\text{Attention}(Q, K, V) = \text{softmax}\left(\frac{QK^\top}{\sqrt{d_k}}\right)V,
\]
where $d_k$ is the dimensionality of the key vectors per attention head.

We set the number of attention heads to 4. Each head independently learns its own Q, K and V projections. Outputs from all heads are concatenated and linearly projected back to the original channel dimension. Layer normalization and dropout are applied after residual connections to stabilize training. 

In our implementation, the feature maps at the bottleneck have 512 channels, which are evenly split across the 4 heads which result in a token embedding size of 128 per head. This dimensionality is preserved across both EG-MHSA layers at the bottleneck.

Importantly, we omit positional encodings in our novel EG-MHSA design. The learned edge priors already encode meaningful spatial structure, simplifying the architecture while maintaining effectiveness in capturing both local and global context.

\subsection{Bottleneck Integration}
The EG-MHSA modules are placed sequentially at the bottleneck, operating directly on the output of the deepest encoder layer. The processing sequence is:

\begin{align}
z &= \text{DoubleConv}(x_4), \nonumber \\
z &= \text{EG\mbox{-}MHSA}_1(z, E), \nonumber \\
z &= \text{EG\mbox{-}MHSA}_2(z, E),
\end{align}

where $x_4$ is the last encoder feature map and $E$ is the edge prior map. The final output $z$ is passed into the decoder as the starting point for upsampling. This structure enables the model to learn both localized edge detail and global context prior to reconstruction.

\subsection{Loss Function}
To supervise training, we adopt a hybrid loss function combining binary cross-entropy (BCE) and Dice loss:
\[
\mathcal{L}_{\text{total}} = \mathcal{L}_{\text{BCE}} + \mathcal{L}_{\text{Dice}}.
\]
The BCE loss penalizes pixel-wise classification errors by measuring the binary cross-entropy between predicted logits and ground truth labels. The Dice loss complements this by maximizing the overlap between predicted and ground truth masks, making it particularly effective for imbalanced segmentation tasks. Together, the combined loss encourages both accurate pixel classification and structural consistency.

\section{Experiments and Results}

This section outlines the experimental setup and presents the results of our study.

We utilize a total of 1,593 H-$\alpha$ observations from the GONG dataset, along with their corresponding MAGFILO annotations. After filtering out observations without annotations, 1,439 images are retained. This ensures that each H-$\alpha$ image in the final dataset contains at least one filament in its ground truth mask.

The dataset was split into 1,295 training samples, 45 validation samples, and 99 test samples. Training was performed for 50 epochs, ensuring that no validation or test samples were used during training to prevent data leakage. The models were trained using the Adam optimizer with a learning rate of $10^{-4}$. All models were trained under identical settings to ensure a fair comparison. Training was conducted end-to-end, without the use of pretrained weights or data augmentation.

To establish a strong foundation for evaluating the effectiveness of our proposed architecture, we implemented three baseline models: U-Net, U-Net + MHSA (without positional encoding), and U-Net + MHSA (with positional encoding). Table~\ref{architecture-comparison} summarizes the key architectural components, modifications, and training configurations used across all model implementations.

To ensure a fair comparison, all models were trained without transfer learning, using an end-to-end approach on the dataset that was preprocessed through the pipeline shown in Fig.~\ref{processing}. Consistent loss functions and optimization settings were applied across all models.

\begin{figure*}[htbp]
  \centering
  \begin{adjustbox}{max width=\textwidth, max height=0.28\textheight, keepaspectratio}
    \includegraphics{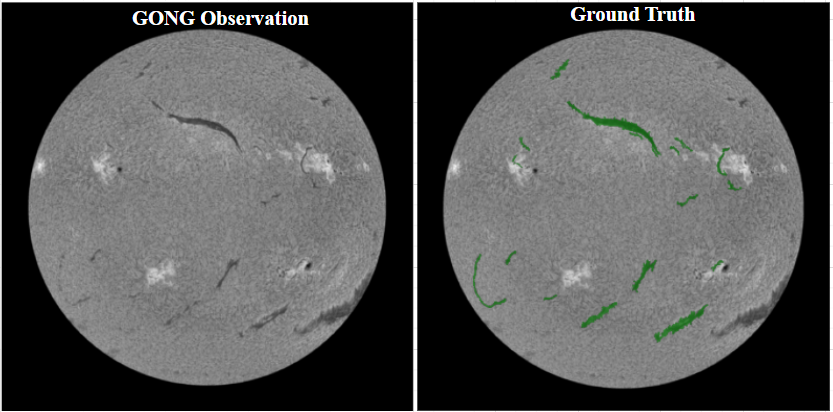}
  \end{adjustbox}
  \vspace{0.5em}

  \begin{adjustbox}{max width=\textwidth, max height=0.28\textheight, keepaspectratio}
    \includegraphics{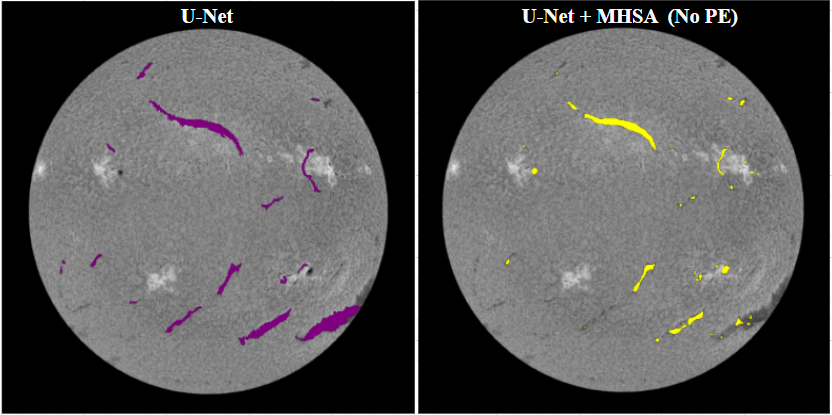}
  \end{adjustbox}
  \vspace{0.5em}

  \begin{adjustbox}{max width=\textwidth, max height=0.28\textheight, keepaspectratio}
    \includegraphics{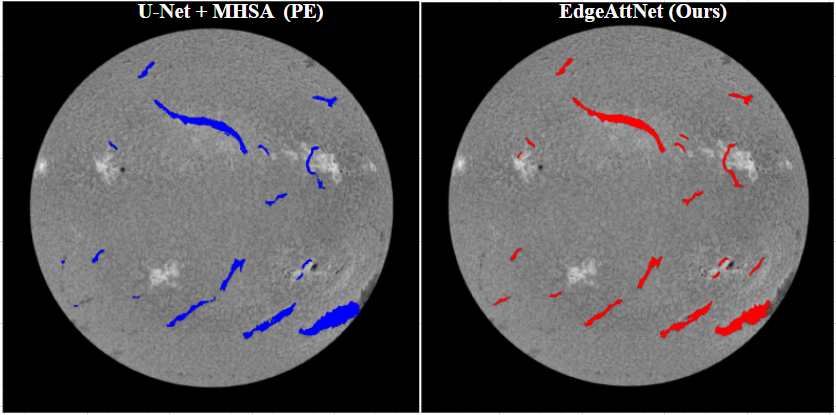}
  \end{adjustbox}

  \caption{Segmentation predictions for July 14, 2022, 18:53:52 — the 10th observation, taken from one of the randomly selected samples shown in Figure~\ref{boxplots}.}
  \label{fig:combined}
\end{figure*}

First, we evaluate each trained model on the test split, with the results summarized in Table~\ref{scoretable}. As shown, EdgeAttNet consistently outperforms the baselines, achieving a $\mathit{mIoU}_{\mathit{pairwise}}$ of 0.6451 and a $\mathit{mIoU}_{\mathit{multiscale}}$ of 0.7032. Here, the ``m'' in $\mathit{mIoU}$ denotes the arithmetic mean, representing the average IoU score across all samples in the test split. These results demonstrate that our method achieves superior segmentation accuracy across all evaluation metrics.

To further assess performance variability and model robustness, we randomly selected 12 H-$\alpha$ observations from the test set and applied all trained models to these samples. This allows us to visualize the distribution of evaluation metrics. Figure~\ref{boxplots} presents boxplots for all baseline models and EdgeAttNet. From these results, we observe that EdgeAttNet achieves the highest $\mathit{mIoU}_{\mathit{pairwise}}$ and $\mathit{mIoU}_{\mathit{multiscale}}$ scores in $\approx 50\%$ of the selected samples and performs comparably to the baselines in the remaining cases. For example, the first sample in Figure~\ref{boxplots} shows an $\mathit{mIoU}_{\mathit{pairwise}}$ of $\approx 0.71$ and an $\mathit{mIoU}_{\mathit{multiscale}}$ of $\approx 0.85$. It is also worth noting that in Figure~\ref{boxplots}, the lowest scores of our proposed method across all randomly selected samples are always better than or comparable to those of the other models. These findings highlight the robustness and consistency of EdgeAttNet in filament segmentation.

\subsection{Comparative Analysis of Models}

We compare the performance of each model by visualizing their predictions against the ground truth mask for the first sample in one of the randomly selected test cases. This particular sample was chosen because its ground truth includes numerous filament annotations, with barbs that are clearly represented. Figure~\ref{fig:combined} shows the predictions alongside the ground truth mask and the predictions of all models. 

As can be seen, our proposed method captures the shape and orientation of filament barbs more accurately than the baseline models and achieves greater overlap with the ground truth regions. Moreover, it avoids incorrectly segmenting sunspots as filaments.






\section{Conclusion}

In this work, we proposed EdgeAttNet, a U-Net based segmentation model enhanced with a novel Edge Guided Multihead Self Attention (EG MHSA) mechanism for detecting solar filaments in H$\alpha$ observations. By integrating learned edge priors directly into the attention module, EdgeAttNet effectively captures fine scale morphological features such as filament barbs and spines, which are critical for scientific interpretation but are often overlooked by conventional architectures.

We trained and evaluated our model on the MAGFILO dataset and implemented a dedicated preprocessing pipeline to mitigate solar imaging artifacts like limb darkening and background interference. EdgeAttNet consistently outperformed strong baselines including U-Net, U-Net with MHSA (with and without positional encodings) across all evaluation metrics, demonstrating its superior ability to capture both global context and local detail while remaining robust.

One of the key advantages of EdgeAttNet is its computational efficiency. It achieves better segmentation performance with significantly fewer parameters: EdgeAttNet contains only 22,658,891 trainable parameters, compared to 31,030,593 in U-Net, 35,231,041 in U-Net with MHSA (without positional encodings), and 35,362,113 in U-Net with MHSA and positional encodings. This reduction in model complexity enables faster training and inference while enhancing generalization, particularly in spatially structured domains like solar imaging.

Furthermore, the use of edge priors eliminates the need for explicit positional encodings, simplifying the architecture without sacrificing spatial awareness. The model’s ability to accurately delineate complex filament structures, especially barbs, makes it well suited for downstream applications such as chirality classification, where understanding the orientation of these features is critical for magnetic field interpretation.

\bibliographystyle{IEEEtran}  
\bibliography{references}     

\begin{thebibliography}{10}
\providecommand{\url}[1]{#1}
\csname url@samestyle\endcsname
\providecommand{\newblock}{\relax}
\providecommand{\bibinfo}[2]{#2}
\providecommand{\BIBentrySTDinterwordspacing}{\spaceskip=0pt\relax}
\providecommand{\BIBentryALTinterwordstretchfactor}{4}
\providecommand{\BIBentryALTinterwordspacing}{\spaceskip=\fontdimen2\font plus
\BIBentryALTinterwordstretchfactor\fontdimen3\font minus \fontdimen4\font\relax}
\providecommand{\BIBforeignlanguage}[2]{{%
\expandafter\ifx\csname l@#1\endcsname\relax
\typeout{** WARNING: IEEEtran.bst: No hyphenation pattern has been}%
\typeout{** loaded for the language `#1'. Using the pattern for}%
\typeout{** the default language instead.}%
\else
\language=\csname l@#1\endcsname
\fi
#2}}
\providecommand{\BIBdecl}{\relax}
\BIBdecl

\bibitem{gibson2018solar}
S.~E. Gibson, ``Solar prominences: theory and models: Fleshing out the magnetic skeleton,'' \emph{Living reviews in solar physics}, vol.~15, no.~1, p.~7, 2018.

\bibitem{eastwood2017scientific}
J.~Eastwood, R.~Nakamura, L.~Turc, L.~Mejnertsen, and M.~Hesse, ``The scientific foundations of forecasting magnetospheric space weather,'' \emph{Space Science Reviews}, vol. 212, pp. 1221--1252, 2017.

\bibitem{martin1998conditions}
S.~F. Martin, ``Conditions for the formation and maintenance of filaments--(invited review),'' \emph{Solar Physics}, vol. 182, no.~1, pp. 107--137, 1998.

\bibitem{martin1998filament}
------, ``Filament chirality: A link between fine-scale and global patterns,'' in \emph{International Astronomical Union Colloquium}, vol. 167.\hskip 1em plus 0.5em minus 0.4em\relax Cambridge University Press, 1998, pp. 419--429.

\bibitem{hao2016can}
Q.~Hao, Y.~Guo, C.~Fang, P.-F. Chen, and W.-D. Cao, ``Can we determine the filament chirality by the filament footpoint location or the barb-bearing?'' \emph{Research in Astronomy and Astrophysics}, vol.~16, no.~1, p. 001, 2016.

\bibitem{ahmadzadeh2019toward}
A.~Ahmadzadeh, S.~S. Mahajan, D.~J. Kempton, R.~A. Angryk, and S.~Ji, ``Toward filament segmentation using deep neural networks,'' in \emph{2019 IEEE International Conference on Big Data (Big Data)}.\hskip 1em plus 0.5em minus 0.4em\relax IEEE, 2019, pp. 4932--4941.

\bibitem{zhu2025flat}
G.~Zhu, G.~Lin, X.~Yang, and C.~Zeng, ``Flat u-net: An efficient ultralightweight model for solar filament segmentation in full-disk h-alpha images,'' \emph{arXiv preprint arXiv:2502.07259}, 2025.

\bibitem{ronneberger2015u}
O.~Ronneberger, P.~Fischer, and T.~Brox, ``U-net: Convolutional networks for biomedical image segmentation,'' in \emph{Medical image computing and computer-assisted intervention--MICCAI 2015: 18th international conference, Munich, Germany, October 5-9, 2015, proceedings, part III 18}.\hskip 1em plus 0.5em minus 0.4em\relax Springer, 2015, pp. 234--241.

\bibitem{petit2021u}
O.~Petit, N.~Thome, C.~Rambour, L.~Themyr, T.~Collins, and L.~Soler, ``U-net transformer: Self and cross attention for medical image segmentation,'' in \emph{International Workshop on Machine Learning in Medical Imaging}.\hskip 1em plus 0.5em minus 0.4em\relax Springer, 2021, pp. 267--276.

\bibitem{qin2020match}
X.~Qin, C.~Wu, H.~Chang, H.~Lu, and X.~Zhang, ``Match feature u-net: Dynamic receptive field networks for biomedical image segmentation,'' \emph{Symmetry}, vol.~12, no.~8, p. 1230, 2020.

\bibitem{oktay2018attention}
O.~Oktay, J.~Schlemper, L.~Le~Folgoc, M.~Lee, M.~Heinrich, K.~Misawa, K.~Mori, S.~McDonagh, N.~Y. Hammerla, B.~Kainz, B.~Glocker, and D.~Rueckert, ``Attention u-net: Learning where to look for the pancreas,'' in \emph{MICCAI}.\hskip 1em plus 0.5em minus 0.4em\relax Springer, 2018, adds attention gates in U-Net decoder to focus on relevant semantic regions.

\bibitem{xie2021segformer}
E.~Xie, W.~Wang, Z.~Yu, A.~Anandkumar, J.~M. Alvarez, and P.~Luo, ``Segformer: Simple and efficient design for semantic segmentation with transformers,'' \emph{Advances in neural information processing systems}, vol.~34, pp. 12\,077--12\,090, 2021.

\bibitem{woo2018cbam}
S.~Woo, J.~Park, J.-Y. Lee, and I.~S. Kweon, ``Cbam: Convolutional block attention module,'' in \emph{ECCV}.\hskip 1em plus 0.5em minus 0.4em\relax Springer, 2018, introduces a lightweight, sequential channel and spatial attention block to enhance CNN feature maps.

\bibitem{hu2018squeeze}
J.~Hu, L.~Shen, and G.~Sun, ``Squeeze-and-excitation networks,'' in \emph{Proceedings of the IEEE/CVF Conference on Computer Vision and Pattern Recognition (CVPR)}, 2018, pp. 7132--7141.

\bibitem{roy2018concurrent}
A.~G. Roy, N.~Navab, and C.~Wachinger, ``Concurrent spatial and channel ‘squeeze \& excitation’in fully convolutional networks,'' in \emph{International conference on medical image computing and computer-assisted intervention}.\hskip 1em plus 0.5em minus 0.4em\relax Springer, 2018, pp. 421--429.

\bibitem{ge2022edgeformer}
T.~Ge, S.-Q. Chen, and F.~Wei, ``Edgeformer: a parameter-efficient transformer for on-device seq2seq generation,'' \emph{arXiv preprint arXiv:2202.07959}, 2022.

\bibitem{ahmadzadeh2024dataset}
A.~Ahmadzadeh, R.~Adhyapak, K.~Chaurasiya, L.~A. Nagubandi, V.~Aparna, P.~C. Martens, A.~Pevtsov, L.~Bertello, A.~Pevtsov, N.~Douglas \emph{et~al.}, ``A dataset of manually annotated filaments from h-alpha observations,'' \emph{Scientific Data}, vol.~11, no.~1, p. 1031, 2024.

\bibitem{harvey1996global}
J.~Harvey, F.~Hill, R.~Hubbard, J.~Kennedy, J.~Leibacher, J.~Pintar, P.~Gilman, R.~Noyes, A.~Title, J.~Toomre \emph{et~al.}, ``The global oscillation network group (gong) project,'' \emph{Science}, vol. 272, no. 5266, pp. 1284--1286, 1996.

\bibitem{wang2014new}
D.~Wang and Y.~Shang, ``A new active labeling method for deep learning,'' in \emph{2014 International joint conference on neural networks (IJCNN)}.\hskip 1em plus 0.5em minus 0.4em\relax IEEE, 2014, pp. 112--119.

\bibitem{men2017background}
A.~Men’shchikov, ``Background derivation and image flattening: getimages,'' \emph{Astronomy \& Astrophysics}, vol. 607, p. A64, 2017.

\bibitem{wink2004denoising}
A.~M. Wink and J.~B. Roerdink, ``Denoising functional mr images: a comparison of wavelet denoising and gaussian smoothing,'' \emph{IEEE transactions on medical imaging}, vol.~23, no.~3, pp. 374--387, 2004.

\bibitem{reza2004realization}
A.~M. Reza, ``Realization of the contrast limited adaptive histogram equalization (clahe) for real-time image enhancement,'' \emph{Journal of VLSI signal processing systems for signal, image and video technology}, vol.~38, no.~1, pp. 35--44, 2004.

\bibitem{duda1972use}
R.~O. Duda and P.~E. Hart, ``Use of the hough transformation to detect lines and curves in pictures,'' \emph{Communications of the ACM}, vol.~15, no.~1, pp. 11--15, 1972.

\bibitem{guo2022solar}
X.~Guo, Y.~Yang, S.~Feng, X.~Bai, B.~Liang, and W.~Dai, ``Solar-filament detection and classification based on deep learning,'' \emph{Solar Physics}, vol. 297, no.~8, p. 104, 2022.

\bibitem{diercke2024universal}
A.~Diercke, R.~Jarolim, C.~Kuckein, S.~J.~G. Manrique, M.~Ziener, A.~M. Veronig, C.~Denker, W.~P{\"o}tzi, T.~Podladchikova, and A.~A. Pevtsov, ``A universal method for solar filament detection from h$\alpha$ observations using semi-supervised deep learning,'' \emph{Astronomy \& Astrophysics}, vol. 686, p. A213, 2024.

\bibitem{Ji2023}
H.~Ji, M.~G. Bobra, X.~Sun, Y.~Liu, and J.~T. Hoeksema, ``A systematic magnetic polarity inversion line data set from sdo/hmi magnetograms,'' \emph{The Astrophysical Journal Supplement Series}, vol. 265, no.~2, p.~40, 2023.

\bibitem{lin2014microsoft}
T.-Y. Lin, M.~Maire, S.~Belongie, J.~Hays, P.~Perona, D.~Ramanan, P.~Doll{\'a}r, and C.~L. Zitnick, ``Microsoft coco: Common objects in context,'' in \emph{European conference on computer vision}.\hskip 1em plus 0.5em minus 0.4em\relax Springer, 2014, pp. 740--755.

\bibitem{rezatofighi2019generalized}
H.~Rezatofighi, N.~Tsoi, J.~Gwak, A.~Sadeghian, I.~Reid, and S.~Savarese, ``Generalized intersection over union: A metric and a loss for bounding box regression,'' in \emph{Proceedings of the IEEE/CVF conference on computer vision and pattern recognition}, 2019, pp. 658--666.

\bibitem{ahmadzadeh2021multiscale}
A.~Ahmadzadeh, D.~J. Kempton, Y.~Chen, and R.~A. Angryk, ``Multiscale iou: A metric for evaluation of salient object detection with fine structures,'' in \emph{2021 IEEE International Conference on Image Processing (ICIP)}.\hskip 1em plus 0.5em minus 0.4em\relax IEEE, 2021, pp. 684--688.

\end{thebibliography}

\end{document}